\def\year{2022}\relax
\documentclass[letterpaper]{article} % DO NOT CHANGE THIS
\usepackage{aaai22}  % DO NOT CHANGE THIS
\usepackage{times}  % DO NOT CHANGE THIS
\usepackage{helvet}  % DO NOT CHANGE THIS
\usepackage{courier}  % DO NOT CHANGE THIS
\usepackage[hyphens]{url}  % DO NOT CHANGE THIS
\usepackage{graphicx} % DO NOT CHANGE THIS
\usepackage{amssymb} 
\usepackage{natbib}  % DO NOT CHANGE THIS AND DO NOT ADD ANY OPTIONS TO IT
\usepackage{caption} % DO NOT CHANGE THIS AND DO NOT ADD ANY OPTIONS TO IT
\DeclareCaptionStyle{ruled}{labelfont=normalfont,labelsep=colon,strut=off} % DO NOT CHANGE THIS
\usepackage{algorithm}
\usepackage{algorithmic}

\usepackage{newfloat}
\usepackage{listings}
\floatstyle{ruled}
\newfloat{listing}{tb}{lst}{}
\floatname{listing}{Listing}
\title{Zero Shot Learning for Predicting Energy Usage of Buildings in Sustainable Design}
\author {
    Arun Zachariah\textsuperscript{\rm 1},
    Praveen Rao\textsuperscript{\rm 1},
    Brian Corn\textsuperscript{\rm 2},
    Dominique Davison\textsuperscript{\rm 2}\\
}
\affiliations {
    \textsuperscript{\rm 1} University of Missouri-Columbia, Columbia, Missouri, USA \\
    \textsuperscript{\rm 2} PlanIT Impact, Kansas City, Missouri, USA \\
    azachariah@mail.missouri.edu, praveen.rao@missouri.edu, bcorn@planitimpact.com, ddavison@planitimpact.com
}

\usepackage{bibentry}
\begin{document}

\maketitle

\begin{abstract}
The 2030 Challenge is aimed at making all new buildings and major renovations carbon neutral by 2030. One of the potential solutions to meet this challenge is through innovative sustainable design strategies. For developing such strategies it is important to understand how the various building factors contribute to energy usage of a building, right at design time. The growth of artificial intelligence (AI) in recent years provides an unprecedented opportunity to advance sustainable design by learning complex relationships between building factors from available data. However, rich training datasets are needed for AI-based solutions to achieve good prediction accuracy. Unfortunately, obtaining training datasets are time consuming and expensive in many real-world applications. Motivated by these reasons, we address the problem of accurately predicting the energy usage of new or unknown building types, i.e., those building types that do not have any training data. We propose a novel approach based on zero-shot learning (ZSL) to solve this problem. Our approach uses side information from building energy modeling experts to predict the closest building types for a given new/unknown building type. We then obtain the predicted energy usage for the $k$-closest building types using the models learned during training and combine the predicted values using a weighted averaging function. We evaluated our approach on a dataset containing five building types generated using BuildSimHub, a popular platform for building energy modeling. Our approach achieved better average accuracy than a regression model (based on XGBoost) trained on the entire dataset of known building types. %without taking into account the building type.
\end{abstract}

\section{Introduction}
In 2020, the U.S. Energy Information Administration reported that buildings amounted to 40\% of the total U.S. energy consumption.\footnote{\url{https://www.eia.gov/totalenergy/data/monthly/}} In addition, residential and non-residential buildings contribute significantly to global energy-related CO\textsubscript{2} emissions.\footnote{\url{https://globalabc.org}} Sustainable housing design can eliminate green house gas emissions. For this reason, The 2030 Challenge\footnote{\url{https://architecture2030.org/2030_challenges/2030-challenge}} was issued by Architecture 2030. The goal of this challenge is to ensure that new buildings and major renovations shall be carbon neutral by 2030. 
%To support this challenge, American Institute of Architects (AIA) has issued the 2030 Commitment\footnote{\url{https://www.aia.org/resources/202041-the-2030-commitment}} to help architecture firms incorporate carbon neutrality as a design standard by 2030. 
Sustainable building design is set to become a standard for the 21$^{st}$ century home\footnote{\url{https://fontanarchitecture.com/sustainable-house-design-21-ideas}}. Predicting the energy consumption of a building is necessary for energy planning, management, and conservation. Modeling the energy usage of a building is a complex problem since there are hundreds of factors/features (e.g. area, location, elevation, occupancy, orientation) that must be considered. Moreover, the structural characteristics of a building also impacts the energy usage further adding to its complexity. 

The growth of AI in recent years provides an unprecedented opportunity to advance sustainable design by learning complex relationships between building factors from available real-world data. However, rich training data is needed for AI-based solutions to achieve good prediction accuracy. Unfortunately, obtaining training data is time consuming and expensive in many real-world applications. Motivated by these reasons, we address \textit{the problem of predicting the energy usage of new building types}, for which no training data exists. We build on the success of ZSL for image classification and adapt it for our purpose. The main contributions of our work are as follows:

\begin{itemize}

\item We propose a new approach using ZSL to predict the energy usage of an unknown building type by first identifying the $k$-closest building types.
\item We then compute a weighted average of the predictions from the known types that were used during training. Our approach can leverage side information required for ZSL from either a building energy modeling expert or using dimensionality reduction techniques.
\item We performed an evaluation of our approach and observed that it achieves better accuracy than a single regression XGBoost model that was trained on the data of known building types. (This model ignores the building type during training as it is unknown during prediction.)
\end{itemize}

\section{Related Work and Motivation} 
\subsection{Building Energy Usage Prediction} 
Techniques for predicting building energy usage can be broadly categorized as engineering methods and data-driven approaches~\cite{10.1016/j.rser.2012.02.049}. Traditional engineering methods primarily use physics-based principles to predict energy consumption. They take into account building construction, operation, equipment usage, and environmental information to calculate precise energy consumption. Zhao et al. \cite{10.1016/j.rser.2012.02.049} provide a nice survey of common techniques for energy prediction of buildings. Numerous software tools have been developed over the years for building energy prediction  such as EnergyPlus\footnote{\url{https://energyplus.net/}}, OpenStudio\footnote{\url{https://www.openstudio.net/}}, and BuildSim\footnote{\url{https://www.buildsim.io}}.

%\footnote{\url{https://www.buildingenergysoftwaretools.com/}}

Data-driven approaches can be further classified into statistical techniques and AI-based methods. Statistical techniques try to correlate the energy consumption with the parameters involved in building design using historic data. Bauer et al. \cite{10.1016/S0378-7788(97)00035-2} used correlation for understanding both heating and cooling loads simultaneously. Dhar et al. \cite{10.1115/1.2888072} used Fourier series to model hourly energy usage in commercial buildings. With advances in AI, artificial neural networks (ANNs) have been used to model non-linear relationships between variables for predicting energy consumption \cite{10.1007/s12273-019-0538-0}. Support vector machines (SVMs) were more accurate when limited training data was available \cite{10.1016/j.rser.2012.02.049,10.1016/j.apenergy.2008.11.035}. Gaussian process regression was used due to its ability to quantify and handle different sources of uncertainty in the data \cite{10.1186/s40327-018-0064-7}. Data clustering, aimed at discovering natural grouping in data, was used in energy consumption analysis to sample representative building types \cite{10.1016/j.enbuild.2015.03.036, 10.1016/j.enbuild.2015.02.017}. More recently, XGBoost has been used in numerous applications for predicting and modeling building energy performance \cite{10.1016/j.enbuild.2018.12.032}. It performs ensemble learning, combining the predictions from multiple base learners, enabling it to model complex relationships between attributes in the data. Chakraborty et al. \cite{10.1080/19401493.2018.1498538} showed that XGBoost performed better than ANN-based energy models. Thus, there is growing interest in using AI-based methods for building energy prediction.

\subsection{Zero-Shot Learning}
In recent years, ZSL~\cite{xian2018zero} has attracted much attention in image classification for predicting classes that were not seen during training. ZSL requires some form of side information, which is used to share information between classes so that the knowledge from the known classes (seen during training) is transferred to unseen classes. Side information can be represented by attributes of classes or embeddings (i.e., low-dimensional representation of high-dimensional feature vectors) in a continuous space. ZSL is very useful in real-world applications when there is lack of training data for certain classes.

Romera-Paredes et al. \cite{10.5555/3045118.3045347} developed a simple ZSL approach, which is of particular interest to us. Essentially, a linear model is learned on the training instances to compute a matrix $W$. Using $W$ and a matrix of signatures $S$ of known classes, a matrix $V$ is computed s.t. $W = V \times S$. At inference time, a new matrix of signatures $S^{'}$ (from the unknown/test classes) is used to compute $W^{'} = V \times S'$. This new linear model $W^{'}$ is used to obtain the predicted classes on the test data. 

\subsection{Motivation} 
The aforementioned techniques proposed to predict building energy usage perform well when provided with precise and detailed inputs describing the numerous factors involved in designing the building. Unfortunately, many of these factors are hard to obtain and audit. Furthermore, the success of AI-based techniques depends on high quality training data, which may be time consuming/expensive to produce. For example, running an energy model for a specific building (e.g., using BuildSimHub) can cost thousands of dollars. We hypothesize that ZSL is a promising approach for predicting the energy usage of new/unknown building types. However, modeling the problem of energy prediction using ZSL is non-trivial. Appropriate side information is needed for ZSL to be successful.

\section{Our Approach}
In this section, we describe our ZSL approach for predicting the energy usage of new or unknown building types, i.e., those building types that do not have any training data. 

Our approach draws inspiration from the ZSL approach proposed by Romera-Paredes et al. \cite{10.5555/3045118.3045347}. For side information, let us assume there exists a matrix $\mathbb{S}$, which is handcrafted by domain experts containing information for all the building types $B=\{b_1, b_2, ...\}$ based on several building energy factors/parameters $P=\{p_1, p_2, ...\}$. Note that $\mathbb{S}$ is a $|P|\times|B|$ matrix. We will assume that some of the building types in $B$ are treated as new/unknown. Let $B_u \subseteq  B$ denote these unknown building types. Let $X$ denote the feature vector of the training data instances of known building types $B-B_u$, and $Y$ denote the target vector that represents the building type for each training data instance. 

During training, we first learn $W$ (e.g., using logistic regression) given $X$ and $Y$. Given a set of known building types $B-B_u$, we construct $S$ from $\mathbb{S}$ by dropping the columns corresponding to building types in $B_u$. We then compute $V$ using $S$ and $W$. For each known building type, we use the training data instances for that building type and construct a predictive model for the building energy usage metrics. This model is then used during inference/energy prediction. Algorithm~\ref{algo_train_model} summarizes the training steps.

\begin{algorithm}[tbh]
\caption{Training the models}
\label{algo_train_model}
\begin{algorithmic}[1]
\REQUIRE $X$: Feature vector of training data; $Y$  target vector containing the building type; $S$: Side information for ZSL; $B_u$: Unknown building types
\STATE Learn $W$ using logistic regression given $X$ and $Y$ 
%$\leftarrow$ Model weights trained given $X$ and $Y$
\STATE Construct $S$ by dropping columns from $\mathbb{S}$ using $B_u$
\STATE Compute $V$ s.t. $W \leftarrow V \times S$
\FOR{$b_i \in B-B_u$}
\STATE Learn predictive models using $X_{b_i}$ for the building energy usage metrics
\ENDFOR
\end{algorithmic}
\end{algorithm}

During inference, our goal is to predict the energy usage of $B_u$. (For simplicity, we will assume only one building type in $B_u$ is used during inference.) Our key idea is to first find the $k$-closest known building types to the given unknown building type $b_i \in B_u$. For computing this efficiently, we propose the following steps, summarized in Algorithm~\ref{algo_predict_usage}: Construct $S^{'}$ from $\mathbb{S}$ by dropping all columns in $\mathbb{S}$ for unknown building types $B_u-\{b_i\}$ (Line~\ref{algo_construct_updated_signature}). Compute $W^{'}$ using $S^{'}$ and $V$ (Line~\ref{algo_construct_updated_W}). Let $X^{'}$ denote the feature vector of the test data instances for $b_i$. We then compute $Y^{'}$ (Line~\ref{algo_construct_updated_Y}). Note that $Y^{'}$ is a matrix of real-valued numbers, wherein each row of $Y'$ corresponds to a test data instance. The columns of $Y'$ correspond to the known building types and $b_i$. Thus, we have a score that associates each test data instance with a building type. We ignore the score for $b_i$ and sort the remaining scores (high to low) to obtain the $k$-closest building types.

\begin{algorithm}[tbh]
\caption{Energy usage prediction using ZSL}
\label{algo_predict_usage}
\begin{algorithmic}[1]
\REQUIRE $X^{'}$: Feature vector of test data; $Y$  building type for each training data instance; $S$: Side information from ZSL; $b_i$: Unknown building type; $k$: number of closest building types to consider
\ENSURE $P$: Predicted energy metrics for $b_i$
\STATE \label{algo_construct_updated_signature} Construct $S^{'}$ by dropping columns from $\mathbb{S}$ corresponding to $B_u-\{b_i\}$
\STATE \label{algo_construct_updated_W} $W^{'}$ $\leftarrow$ $V$ $\times$ $S^{'}$
\STATE \label{algo_construct_updated_Y} $Y^{'}$ $\leftarrow$ $X^{'}$ $\times$ $W^{'}$
\FOR{$t$ in $X^{'}$}
\STATE \label{algo_predict_start} Let $(s_1, s_2, ..., s_k)$ denote the sorted scores from $Y^{'}$ for $t$ after ignoring the score for $b_i$
%\STATE $p$ $\leftarrow$ $0$, $E$ $\leftarrow$ $0$
\STATE Let $(e_1, e_2, ...,  e_k)$ denote the predicted values (given $t$) for the $k$ closest building types using the trained regression models 
\STATE Let $P$ denote weighted average of $(e_1, ..., e_k)$ using $(s_1, ..., s_k)$
\ENDFOR \label{algo_predict_end}
% \STATE $p_t$ = $p/k$
%\STATE $P[t]$ = $p/E$
%\ENDFOR\label{algo_predict_end}
\RETURN $P$ 
\end{algorithmic}
\end{algorithm}

% \FOR{$j$ $\leftarrow$ $0$ to $k$}
% \STATE $e_j$ $\leftarrow$ $Y^{'}[j]$ building predictive model output for for input $t$ 
% \STATE $p$ $\leftarrow$ $p$ + \{$e_j$ $\times$ $Normalized(s_j)\}$
% \STATE $E$ $\leftarrow$ $E$ + $e_j$

For a test data instance with feature vector $t$, let $(s_1, s_2, ..., s_k)$ denote the scores for the $k$-closest building types, where $k>0$. We use the predictive models for each of the $k$-closest building types to predict a building energy metric of interest. Let $(e_1, e_2, ...,  e_k)$ denote the predicted values for the unknown building $b_i$. We compute the weighted average of the predicted values, where the weights are computed by either applying \textit{softmax} or any normalization technique on the scores. These steps are shown in Lines~\ref{algo_predict_start}-\ref{algo_predict_end} of Algorithm~\ref{algo_predict_usage}.

If $\mathbb{S}$ is not available from a domain expert, we could employ singular value decomposition (SVD) on simulation data for a given building type. (Tools like BuildSimHub could be used to generate simulation data.) The singular values obtained from SVD can now be used to represent a column of $\mathbb{S}$ for that building type. In summary, robust side information is required for our ZSL approach to succeed.

\section{Experimental Evaluation}
% Table 1 - Start
\begin{table*}[tbh]
\centering
\begin{tabular}{|c|c|c|c|c|c|c|c|c|c|c||c|c|c|}
    \hline
\textbf{\small{Unknown}} & \textbf{\small{Total}} & \multicolumn{3}{c|}{\textbf{\small{TGAS}}} & \multicolumn{3}{c|}{\textbf{\small{COOL}}} & \multicolumn{3}{c||}{\textbf{\small{PFAC}}} & \multicolumn{3}{c|}{\textbf{\small{Avg. Accuracy}}}\\
    \cline{3-14}
\textbf{\small{building}}    & \textbf{\small{\# of}} & \textbf{Base} & \textbf{ZSL$_d$} & \textbf{ZSL$_s$} & \textbf{Base} & \textbf{ZSL$_d$} & \textbf{ZSL$_s$} & \textbf{Base} & \textbf{ZSL$_d$} & \textbf{ZSL$_s$}  & \textbf{Base} & \textbf{ZSL$_d$} & \textbf{ZSL$_s$}\\
\textbf{\small{type}}    & \textbf{\small{records}} & \textbf{line} & \textbf{} & \textbf{} & \textbf{line} & \textbf{} & \textbf{} & \textbf{line} & \textbf{} & \textbf{} & \textbf{line} & \textbf{} & \textbf{}\\
    & & \textbf{(\%)} & \textbf{(\%)} & \textbf{(\%)} & \textbf{(\%)} & \textbf{(\%)} & \textbf{(\%)} & \textbf{(\%)} & \textbf{(\%)} & \textbf{(\%)} & \textbf{(\%)} & \textbf{(\%)} & \textbf{(\%)}\\
    \hline
    \hline
    ED & 13,645 & 56.05 & \textbf{68.68} & 68.11 & \textbf{85.71} & 81.41 & 81.41 & 83.61 & \textbf{88.00} & \textbf{88.00} & 75.13 & \textbf{79.37} & 79.17\\
    \hline
    MU & 25,801 & 71.06 & 78.24 & \textbf{78.25} & 55.58 & \textbf{59.55} & \textbf{59.55} & 74.70 & 78.06 & \textbf{78.46} & 67.11 & 71.95 & \textbf{72.09}\\ 
    \hline
    OF & 13,519 & 84.88 & \textbf{88.12} & \textbf{88.12} & 63.69 & \textbf{84.55} & \textbf{84.55} & \textbf{82.18} & 78.87 & 78.87 & 76.92 & \textbf{83.85} & \textbf{83.85}\\ 
    \hline
    RS & 13,428 & 91.31 & \textbf{92.74} & \textbf{92.74} & 85.82 & \textbf{87.66} & \textbf{87.66} & 87.85 & \textbf{94.30} & \textbf{94.30} & 88.33 & \textbf{91.57} & \textbf{91.57}\\ 
    \hline
    RL & 13,816 & \textbf{87.56} & 85.56 & 85.56 & \textbf{77.15} & 74.76 & 77.14 & 80.51 & 78.73 & \textbf{83.99} & 81.74 & 79.69 & \textbf{82.23}\\
    \hline
\end{tabular}
\caption{Evaluation results (best results shown in bold)}
\label{table1}
\end{table*}
% Table 1 - End

\subsection{Dataset and Predicted Energy Metrics}
For our evaluation, we used BuildSimHub\footnote{\url{https://www.buildsim.io}} to generate a dataset by using the Monte Carlo method through repeated sampling of the various factors/parameters involved in the design of a building type. The simulations were done for five building types, namely, \textit{Educational} (ED), \textit{MixedUse} (MU), \textit{Office} (OF), \textit{Retail-Standalone} (RS), and \textit{Retail-Stripmall} (RL). The dataset had 69 features that contained both categorical and continuous variables. The total instances/records for each building type are shown in Table \ref{table1}.

We predicted three metrics of interest for a given building type. These included \textit{Total Site Gas Energy Usage Intensity} (TGAS), \textit{Cooling Electricity Demand} (COOL) and \textit{Facility Peak Electricity Demand} (PFAC). TGAS denotes the total gas usage of a building per square foot per year. COOL denotes the amount of energy required for cooling in a given duration of time. PFAC denotes the highest amount of electricity consumed in a given duration of time.

%\footnote{\url{https://www.cloudlab.us/}}

\subsection{Setup and Implementation}
All the experiments were conducted on a machine in CloudLab~\cite{10.5555/3358807.3358809}. We used Python (v3.6), NumPy (v1.19.5), and Pandas (v1.1.5). We also used the Python package for XGBoost\footnote{\url{https://github.com/dmlc/xgboost}} (v1.0.1) and scikit-learn (v0.24.2) for implementing logistic regression (in order to compute the regression coefficients required by our ZSL approach). For normalization, we used the softmax function, included with SciPy (v1.5.4), as it gave us the best results. 

\subsection{Evaluation}
We compared our ZSL-based approach, referred hereinafter as $ZSL$, with a baseline model (Baseline) that used XGBoost. The dataset for each building type was split randomly into train and test instances. (The train-test ratio was set at 9:1.) In order to compare the accuracy of $ZSL$ with Baseline, we performed the following steps: (1) We selected one building type as unknown (e.g., ED) and assumed the remaining were known building types for training (e.g., MU, OF, RS, and RL). (2) A predictive model was trained on the training sets of the known building types. (3) For the unknown building type, we used each data record in the test set that denotes specific building parameter values, and predicted the three metrics, namely, TGAS, COOL, and PFAC. (4) Finally, we computed the accuracy of the predictions for the three metrics of the unknown building type. We report the average accuracy (\%) across all the test instances of the unknown building type. We repeated the above steps for all the five building types. 

For Baseline, the predictive model was trained using XGBoost. During training, the building types of the known buildings were ignored as the building type of the test data was assumed to be unknown. We performed hyper-parameter tuning (on maximum tree depth and rate of learning) and $k$ fold cross validation by splitting the training instances into 5 folds. (We observed that our XGBoost models outperformed neural networks for regression on our dataset. Hence, we do not use neural networks for comparison.)

For $ZSL$, the predictive model composed of 4 XGBoost models trained separately on each known building type as we first identify the closest building types during prediction on an unknown building type. We evaluated two variations of $ZSL$ based on how the signature matrix was constructed. The first one (denoted by $ZSL_d$) used the signature matrix provided by a domain expert. The second one (denoted by $ZSL_s$) used the signature matrix generated by applying SVD on the simulated data of a building type.

%(Other regression models can also be used.) 

The evaluation results are shown in Table~\ref{table1}. Out of 15 cases, $ZSL_d$ or $ZSL_s$ achieved the best accuracy for 12 cases compared to Baseline. (The best results for $ZSL_d$ and $ZSL_s$ were obtained for $k=4$.) We also computed the average of the accuracy (\%) for the three metrics for each unknown building type. Once again, our ZSL-based approaches was \textit{the winner in all cases}. These results demonstrate the benefit of using ZSL for predicting the energy usage of unknown building types. Interestingly, the signature matrices used by $ZSL_d$ and $ZSL_s$, though independently generated, were both very useful side-information for prediction.

\section{Conclusion}
We proposed a novel approach based on ZSL for predicting the energy usage of an unknown building type. Our approach predicts the $k$-closest known building types by using side information from building energy modeling experts. The predicted energy usage of a new/unknown building type is the weighted average of the predictions from these $k$-closest building types. We evaluated our approach on a simulated dataset created using BuildSimHub for five building types to predict three different energy usage metrics. Our approach achieved better average accuracy compared to XGBoost-based models trained without taking into account the building type. Our work aims to advance the field of sustainable design using AI-based techniques and positively impact multiple engineering disciplines.

\paragraph{Acknowledgments:} This work was supported by the National Science Foundation under Grant No. 1747751.

\footnotesize
\bibliographystyle{aaai22}
\bibliography{aaai22}

\end{document}